\documentclass{ws-procs11x85}

\ifdefined\pdfpagewidth
\fi

\usepackage{ws-procs-thm}
\usepackage{amsmath}
\usepackage{amssymb}
\usepackage{booktabs}
\usepackage{graphicx}
\usepackage{microtype}
\usepackage{url}
\usepackage{hyperref}
\hypersetup{hidelinks,pageanchor=false}
\newcommand{\NUnits}{192}
\newcommand{\NBoot}{10000}

\newcommand{\NSelectFitUnits}{386}
\newcommand{\NSelectPickUnits}{384}

\newcommand{\NTabWires}{645}
\newcommand{\NTabTables}{5{,}152}

\newcommand{\NTabCellsEmpty}{825}
\newcommand{\TabCellsEmptyPct}{1.00}

\newcommand{\TabRidge}{0.03}
\newcommand{\TabCap}{32}

\newcommand{\NTabUsedFullFold}{4{,}797}

\newcommand{\TabGateRadius}{18}
\newcommand{\TabGateWeight}{1}
\newcommand{\TabOperatingQ}{0.09}

\newcommand{\TabRidgeLow}{0.8045}
\newcommand{\TabRidgeLowValue}{0.03}
\newcommand{\TabRidgeMid}{0.8042}
\newcommand{\TabRidgeMidValue}{0.1}
\newcommand{\TabRidgeHigh}{0.8023}
\newcommand{\TabRidgeHighValue}{0.3}

\newcommand{\NSplitsCv}{4}

\newcommand{\NSplitsRh}{25}

\newcommand{\PreRegTestDelta}{+0.0281}
\newcommand{\PreRegTestCI}{[+0.0117, +0.0427]}

\newcommand{\MatchPTwoQ}{0.09}

\newcommand{\MatchOurQ}{0.09}

\newcommand{\MatchNativeDelta}{+0.0315}

\newcommand{\MatchBOurs}{0.3334}
\newcommand{\MatchBPTwo}{0.3194}
\newcommand{\MatchBDelta}{+0.0140}
\newcommand{\MatchBCI}{[-0.0085, +0.0364]}
\newcommand{\MatchBP}{0.224}

\newcommand{\FullBudgetMCCDelta}{+0.0169}
\newcommand{\FullBudgetMCCCI}{[-0.0092, +0.0424]}

\newcommand{\FullCallRatio}{1.65}

\newcommand{\AudGeomMargin}{+15.2}
\newcommand{\AudGateMargin}{+4.4}
\newcommand{\AudGapToFP}{1.3}

\newcommand{\FairMachine}{Apple M4}
\newcommand{\FairWarmOurs}{0.298}

\newcommand{\FairWarmPTwo}{0.162}
\newcommand{\FairFeatures}{0.212}
\newcommand{\FairScoring}{0.0873}
\newcommand{\FairFeatPct}{71}
\newcommand{\FairRSS}{1180}

\newcommand{\FairColdRatio}{5.98}

\newcommand{\FairModelOurs}{1.79}
\newcommand{\FairModelOursGz}{0.39}
\newcommand{\FairModelPTwo}{22.01}
\newcommand{\FairWarmRatio}{1.84}

\newcommand{\FairCpuRatio}{2.95}

\newcommand{\EndReadsBefore}{4}
\newcommand{\EndLineageBefore}{3}

\newcommand{\EndWins}{115}
\newcommand{\EndLosses}{77}
\newcommand{\EndMedianDelta}{+0.0412}

\newcommand{\PlmFidelity}{0.9987}

\newcommand{\PlmLevel}{0.8235}
\newcommand{\PlmAucDelta}{-0.0243}
\newcommand{\PlmAucCI}{[-0.0465, -0.0033]}

\newcommand{\PlmLoss}{111}

\newcommand{\PlmBudgetFDelta}{-0.0522}
\newcommand{\PlmBudgetFCI}{[-0.0786, -0.0274]}

\newcommand{\PlmBudgetMDelta}{-0.0561}
\newcommand{\PlmBudgetMCI}{[-0.0863, -0.0269]}

\newcommand{\PmLevel}{0.7523}

\newcommand{\PmAucOurs}{+0.0468}
\newcommand{\PmAucCIOurs}{[+0.0251, +0.0682]}

\newcommand{\PmAucPtwoR}{+0.0410}
\newcommand{\PmAucCIPtwoR}{[+0.0211, +0.0614]}

\newcommand{\PmSelfAuc}{0.8677}
\newcommand{\PmSelfPublished}{0.87}

\newcommand{\BcN}{45}
\newcommand{\BcLevelGeo}{0.7980}
\newcommand{\BcLevelOurs}{0.7917}
\newcommand{\BcLevelPtwoR}{0.7523}
\newcommand{\BcLevelPm}{0.7136}
\newcommand{\BcLevelPlm}{0.8672}
\newcommand{\BcOurs}{+0.0063}
\newcommand{\BcOursCI}{[-0.0255, +0.0385]}

\newcommand{\BcOursWin}{27}
\newcommand{\BcOursLoss}{18}
\newcommand{\BcOursPred}{+0.0121}

\newcommand{\BcPlm}{-0.0692}
\newcommand{\BcPlmCI}{[-0.1163, -0.0257]}
\newcommand{\BcPlmBonf}{[-0.1312, -0.0141]}

\newcommand{\BcPtwoR}{+0.0457}
\newcommand{\BcPtwoRCI}{[-0.0100, +0.0935]}
\newcommand{\BcPtwoRBonf}{[-0.0264, +0.1041]}

\newcommand{\BcPm}{+0.0844}
\newcommand{\BcPmCI}{[+0.0384, +0.1271]}
\newcommand{\BcPmBonf}{[+0.0245, +0.1384]}

\newcommand{\ExtN}{57}
\newcommand{\ExtNPositives}{905}

\newcommand{\ExtCutoff}{2024-05-08}
\newcommand{\ExtLevelOurs}{0.8411}
\newcommand{\ExtLevelPtwoR}{0.7968}
\newcommand{\ExtLevelPlm}{0.8751}
\newcommand{\ExtLevelPm}{0.7734}
\newcommand{\ExtAucPtwoR}{+0.0443}
\newcommand{\ExtAucCIPtwoR}{[+0.0162, +0.0724]}
\newcommand{\ExtAucBonfPtwoR}{[+0.0102, +0.0792]}
\newcommand{\ExtWinPtwoR}{42}

\newcommand{\ExtAucPlm}{-0.0340}
\newcommand{\ExtAucCIPlm}{[-0.0701, -0.0006]}
\newcommand{\ExtAucBonfPlm}{[-0.0784, +0.0071]}

\newcommand{\ExtAucPm}{+0.0677}
\newcommand{\ExtAucCIPm}{[+0.0328, +0.1013]}

\newcommand{\ExtQPct}{9}

\newcommand{\ExtBudgetFDelta}{+0.0246}
\newcommand{\ExtBudgetFCI}{[-0.0224, +0.0708]}

\newcommand{\ExtAbstain}{1}

\newcommand{\CvN}{39}
\newcommand{\CvLowPct}{2}
\newcommand{\CvHighPct}{40}

\newcommand{\CvPtwoRFExcl}{10}

\newcommand{\RdLogitAuc}{0.8008}

\newcommand{\RdLogitDelta}{+0.0038}
\newcommand{\RdLogitCI}{[-0.0057, +0.0125]}

\newcommand{\NTrainUnits}{770}

\newcommand{\TabAuc}{0.799}
\newcommand{\TabAucCI}{[0.766, 0.830]}

\newcommand{\PtwoRAuc}{0.793}

\newcommand{\GeoAuc}{0.665}

\newcommand{\TabAucD}{+0.0058}
\newcommand{\TabAucDCI}{[-0.0168, +0.0276]}
\newcommand{\TabAucP}{0.598}

\newcommand{\TabPr}{0.376}
\newcommand{\TabPrCI}{[0.339, 0.411]}

\newcommand{\PtwoRPr}{0.358}

\newcommand{\GeoPr}{0.276}

\newcommand{\TabPrD}{+0.0177}
\newcommand{\TabPrDCI}{[-0.0094, +0.0450]}

\newcommand{\TabMcc}{0.304}
\newcommand{\TabMccCI}{[0.270, 0.338]}

\newcommand{\PtwoRMcc}{0.285}

\newcommand{\GeoMcc}{0.239}

\newcommand{\TabFOne}{0.333}
\newcommand{\TabFOneCI}{[0.301, 0.365]}

\newcommand{\PtwoRFOne}{0.302}

\newcommand{\GeoFOne}{0.287}

\newcommand{\FamChem}{+0.0002}

\newcommand{\FamChemCI}{[-0.0007, +0.0010]}

\newcommand{\FamBkb}{+0.0044}

\newcommand{\FamBkbCI}{[+0.0033, +0.0055]}

\newcommand{\FamSide}{+0.0048}

\newcommand{\FamSideCI}{[+0.0032, +0.0063]}

\newcommand{\FamConf}{+0.0072}

\newcommand{\FamConfCI}{[+0.0055, +0.0089]}

\newcommand{\FamVoid}{+0.0026}

\newcommand{\FamVoidCI}{[+0.0012, +0.0039]}

\newcommand{\FamGeom}{+0.0095}

\newcommand{\FamGeomCI}{[+0.0081, +0.0109]}

\newcommand{\FamDisp}{+0.0070}

\newcommand{\FamDispCI}{[+0.0050, +0.0090]}

\newcommand{\FamDispB}{+0.0063}

\newcommand{\FamDispBCI}{[+0.0049, +0.0077]}

\newcommand{\FamDispAlt}{+0.0005}

\newcommand{\FamDispAltCI}{[-0.0007, +0.0017]}

\newcommand{\FamAddOf}{12}

\begin{document}

\title{Auditable Algebraic Counting Field for Cryptic-Pocket Detection
from Apo Structures}

\author{Shan Yu$^1$ and Xuening Wu$^{2,\dagger}$}

\address{$^1$Independent Researcher, Hangzhou, China\\
$^2$Pfizer, Shanghai, China\\
$^\dagger$E-mail: xuening.wu@pfizer.com}

\begin{abstract}
Cryptic ligand-binding pockets are not apparent in experimentally determined
apo structures, making them difficult to identify from unbound receptor
geometry. A complementary
challenge is to make the structural measurements and learned evidence behind
each prediction directly inspectable. We introduce a supervised algebraic
counting field (ACF) for predicting cryptic-pocket residues from apo structures.
ACF compiles explicit geometric, physicochemical, and topological features into
compact, integer-weighted lookup tables. Each prediction score can be
reconstructed from feature values, training counts, table weights, and spatial
aggregation, without sequence search, structural-template transfer, or a
protein language model at inference.
We evaluate ACF on CryptoBench and two locked external collections, separating
ranking performance from the effects of residue-calling budgets. On an external
set of 57 post-CryptoBench apo--holo units, ACF exceeded P2Rank by $+0.044$ in
mean paired ROC-AUC (multiplicity-adjusted 95\% CI [$+0.010$, $+0.079$]).
The advantage was dataset-dependent: official-fold ROC-AUC and matched-budget
F1 differences against P2Rank remained unresolved, and a second external
evaluation did not confirm gains from added structural features. ACF thus
provides a compact predictor with externally validated signal and an
inspectable path from structural measurements and training counts to residue
scores.
\end{abstract}


\copyrightinfo{Accepted for publication in Pacific Symposium on Biocomputing
(PSB) 2027.}

\section{Introduction}

Cryptic ligand-binding sites may be absent or closed in an observed apo
structure, although local geometry and chemistry may indicate where opening
is possible.\cite{vajda2018cryptic,cimermancic2016cryptosite,
beglov2018origins,bowman2012fluctuations} Identifying their lining residues
therefore tests structure-based functional inference beyond annotation
transfer from homologous sequences or structural templates.

Structure-based predictors span explicit geometry, energetic probes,
simulation, and learned representations. The geometry-based method Fpocket detects cavities from the spatial arrangement of protein atoms.\cite{leguilloux2009fpocket} Other
established approaches include geometric and physicochemical site scoring in
SiteMap and DoGSiteScorer and energetic probe mapping in FTMap.
\cite{halgren2009sitemap,volkamer2012dogsite,kozakov2015ftmap} DeepSite,
Kalasanty, and DeepPocket instead formulate pocket detection or segmentation
with three-dimensional convolutional networks.
\cite{jimenez2017deepsite,stepniewska2020kalasanty,
aggarwal2022deeppocket}
P2Rank learns ligandability scores for local chemical neighborhoods on the
protein surface.\cite{krivak2018p2rank}
PocketMiner uses a graph neural network to identify residues likely to
participate in cryptic-pocket opening.\cite{meller2023pocketminer} CryptoBench
adds a supervised protein-language-model baseline (pLM-NN), drawing on the
sequence representations learned by large protein language models, and
sequence-clustered splits for cryptic-site evaluation.
\cite{rives2021proteinlanguage,skrhak2025cryptobench} These methods show that
single structures and learned representations contain useful site information.
They also motivate a complementary question: can a predictor remain competitive
while allowing each score to be mechanically reconstructed from explicit local
measurements and training counts? This focus on traceability distinguishes
explanations constructed after prediction from models whose predictive
computations are directly inspectable.\cite{rudin2019interpretable}

We answer this question with an \emph{algebraic counting field} (ACF)---a
discrete counting architecture over local structural neighborhoods in which
every score decomposes into explicit table lookups and integer-weighted
additions. Within each protein chain, structural features are rank-discretized
into four levels and paired to address compact lookup tables compiled from
training-fold residue counts. A regularized compilation step assigns bounded
integer multiplicities to the tables. The deployed score requires no
neural-network forward pass or external database search, although the method is
\emph{not} training-free: its table counts, residue propensity, and integer
multiplicities are derived from the training fold.

\begin{figure}[tbh]
\centering
\includegraphics[width=\textwidth]{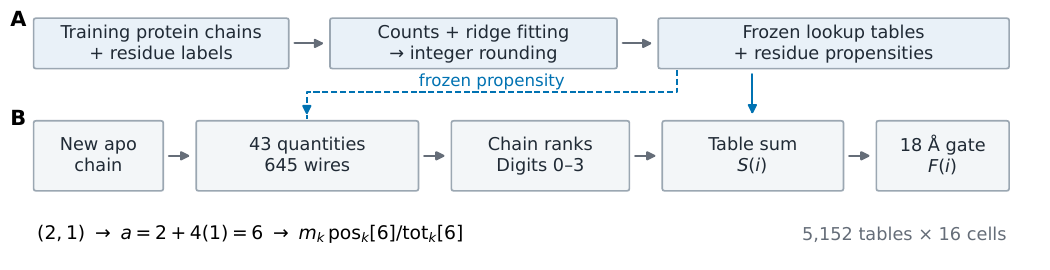}
\caption{ACF workflow. (A) Training uses apo protein chains and residue labels to compile lookup-table counts, integer weights, and residue-type propensities, which are then frozen. (B) Prediction uses a new apo chain's structural features and within-chain ranks with the frozen parameters to obtain residue scores through table lookup and spatial aggregation; no residue labels are required.}
\label{fig:workflow}
\end{figure}

The contribution is an explicit structural representation with a compact,
exactly decomposable readout, evaluated against strong baselines with ranking
and operating-point effects separated. Two external reads support transferable
signal but delimit accuracy claims; controlled training-fold probes identify
additional structural information whose benefit was not confirmed externally.
Figure~\ref{fig:workflow} summarizes the training and prediction workflow.

\section{Materials and Methods}

\subsection{Benchmark, evaluation units, and leakage controls}

We use the official CryptoBench apo split, in which the training and test
partitions are disjoint under MMseqs2 clustering at 10\% sequence
identity.\cite{skrhak2025cryptobench,steinegger2017mmseqs2}
Receptor structures are retrieved from the RCSB Protein Data Bank
(RCSB PDB),\cite{burley2019rcsb} and every coordinate and label file is
verified against a recorded SHA-256 hash. Data loading halts if the manifest is
absent, a file hash has changed, the declared clustering metadata differ from
the recorded values, or a cluster identifier occurs in both partitions.

Each evaluation unit is identified by a PDB entry--chain pair. The official
analysis includes \NUnits{} single-chain units drawn from 190 PDB entries.
Thirty-eight cases requiring multi-chain residue mapping, plus one chain
not retrievable in legacy PDB format, are excluded. The multi-chain cases are
excluded because
the benchmark labels do not unambiguously assign residues to individual
chains. This exclusion is an explicit scope restriction. All structure-based
predictors receive only apo-receptor coordinates. A runtime guard rejects
inputs containing non-solvent heteroatoms or other non-polymer atoms,
preventing ligand coordinates from entering receptor-only predictions. Table~\ref{tab:partitions} summarizes the frozen evaluation partitions.

\begin{table}[tbh]
\tbl{Frozen CryptoBench partitions. Each unit is a PDB entry--chain pair.}
{\begin{minipage}{\textwidth}
\centering
\begin{tabular*}{0.85\textwidth}
{@{\extracolsep{\fill}}lrrl@{}}
\toprule
Partition & Units & Residues & Role\\
\midrule
Training fold
  & \NTrainUnits{} & 234,838 & ACF compilation\\
\quad Fit half
  & \NSelectFitUnits{} & 118,716 & Candidate-readout fitting\\
\quad Selection half
  & \NSelectPickUnits{} & 116,122 & Configuration selection\\
Official test subset
  & \NUnits{} & \multicolumn{1}{c}{---} & Final paired evaluation\\
\bottomrule
\end{tabular*}
\end{minipage}}
\label{tab:partitions}
\end{table}

\subsection{Local quantities, wires, and ACF compilation}

Each residue has 43 explicit base quantities: 35 geometric and topological
descriptors, seven fixed physicochemical constants, and one Laplace-smoothed
residue-type propensity estimated only from the training fold.
The structural descriptors cover surface exposure, local spectral
geometry, density-field calculus, ultrametric organization, curvature,
anisotropy, and global position. The seven fixed physicochemical quantities
are Kyte--Doolittle hydropathy,\cite{kyte1982hydropathy} Zamyatnin side-chain
volume,\cite{zamyatnin1972volume} aromatic-ring count, formal charge,
hydrogen-bond donor and acceptor counts, and the number of rotatable
side-chain bonds. The public implementation documents all definitions, sources, neighborhoods,
boundary cases, and expansion rules. Geometric and physicochemical extensions
require recompilation and separate evaluation. Propensity is fixed during
prediction and re-estimated for a different training fold.

For residue \(i\), define \mbox{\(N_r(i)=\{j:\lVert c_i-c_j\rVert\le r\}\)}.
Here, \(c_i\) is the centroid of its observed heavy atoms, and the intra-chain
neighborhood includes \(i\). Each quantity \(x\) is used directly and through
its neighborhood mean, standard deviation, centered difference, and local rank:
\[
\mathrm{sd}_r[x]_i =
\sqrt{\max\!\left(
  \overline{x^2}_{N_r(i)}
  - \bigl(\overline{x}_{N_r(i)}\bigr)^2,0\right)
},
\quad
D_r[x]_i =
x_i-\overline{x}_{N_r(i)},
\quad
K_r[x]_i =
\frac{
  \left|\{j\in N_r(i):x_j<x_i\}\right|
}{
  |N_r(i)|
}.
\]
Each resulting residue-level scalar feature is called a wire. Neighborhood
means are evaluated at 6, 10, 14, 20, and 26\,\AA, whereas dispersion,
centered difference, and local rank are evaluated at 6, 14, and 20\,\AA.
These predefined transformations retain all 15 versions of each base
quantity, yielding \(43(1+5+3+3+3)=\NTabWires{}\) wires without a further
feature-selection step.

Each wire is discretized separately within the scored chain. For a chain of
\(n\) residues, let \(r_j(i)\) be the zero-based rank of wire \(j\), with ties
assigned their mean rank. Its digit is
\(d_j(i)=\min\{3,\lfloor4r_j(i)/\max(n-1,1)\rfloor\}\).
This discretization is invariant to strictly increasing transformations of
that wire; ties need not
produce equally populated bins. No training-set bin edges are transferred.

A table's \emph{width} is the number of input wires. For a width-2 table
with ordered wire indices \((u_k,v_k)\), the address is
\(a_k(i)=d_{u_k}(i)+4d_{v_k}(i)\in\{0,\ldots,15\}\).
Thus a table has \(4^2=16\) cells: for example, digits \((2,1)\) select cell 6.
The bank is generated by 16 successive permutations of all 645 wire indices
using NumPy \texttt{default\_rng(20260725)}. Adjacent indices are paired and
the final singleton is discarded in each round; pairs are not selected by
labels or biological criteria. Each round contributes \(\lfloor645/2\rfloor=322\)
tables, hence \(16\times322=5{,}152\) tables and
\(5{,}152\times16=82{,}432\) cells. A wire is used at most once per round;
partners can recur across rounds. With the frozen seed, 629 wires occur in
16 tables and 16 wires in 15 tables, so none is absent from the complete bank.

For table \(k\), cell \(a\) stores two training-fold counts:
\(\mathrm{tot}_k[a]\), the number of residues reaching it, and
\(\mathrm{pos}_k[a]\), the number labeled cryptic. Its value is the unsmoothed
fraction \(\mathrm{pos}_k[a]/\mathrm{tot}_k[a]\), or the training-fold base
rate for an unseen cell. There are \NTabCellsEmpty{} unseen cells
(\TabCellsEmptyPct\%) in the full-fold compilation.

The unaggregated residue score is
\[
S(i)=
\sum_{k=1}^{\NTabTables}
m_k
\frac{\mathrm{pos}_k[a_k(i)]}{\mathrm{tot}_k[a_k(i)]},
\qquad
m_k\in\mathbb{Z}\cap[-\TabCap{},\TabCap{}],
\]
with the training-fold base rate substituted for an unseen cell. The deployed
score is
\[
F(i)=S(i)+\overline{S}_{18}(i)
\frac{\mathrm{sd}(S)}
{\mathrm{sd}(\overline{S}_{18})}.
\]
Here, \(\overline{S}_{\TabGateRadius}(i)\) is the mean score among residues
within \TabGateRadius\,\AA\ of residue \(i\), and the standard deviations are
computed across residues in the scored chain. This spatial gate favors
coherent clusters of high-scoring residues over isolated high scores. The
standard-deviation ratio places the neighborhood term on the same dispersion
scale as \(S\). Radius \TabGateRadius{}\,\AA\ and weight \TabGateWeight{}
were selected on a UniProt accession-disjoint training half and frozen;
this internal split does not guarantee sequence-cluster separation.
The gate is skipped if the neighborhood mean has zero standard deviation.

The integer multiplicities \(m_k\) are obtained by solving a
ridge-regularized closed-form system over training-fold table outputs and
rounding the resulting coefficients. Let \(\Sigma\) be the within-class scatter of table outputs divided by
\(\max(N-2,1)\), where \(N\) is the number of training residues, and let
\(\mu_1,\mu_0\) be the class means. The solution is
\[
w=\left[\Sigma+\left(\TabRidge\,\frac{\operatorname{tr}\Sigma}{K}
+10^{-12}\right)I\right]^{-1}(\mu_1-\mu_0),\qquad K=5{,}152.
\]
Weights are \(m_k=\operatorname{round}(\TabCap\,w_k/\max_j|w_j|)\),
or all zero when the solution is zero. The ridge penalty scales with the mean diagonal,
not the unnormalized trace. Compilation uses a real-valued solve; inference
uses floating-point cell fractions, integer coefficients, and spatial statistics. Of the
\NTabTables{} compiled tables, \NTabUsedFullFold{} have nonzero multiplicity
in the full-training-fold ACF.

\subsection{Baselines and operating points}

\emph{Rigid geometry} (\texttt{geometric\_foundation}) is a training-free
static-coordinate baseline that ranks free-space points by ray enclosure and
a local cavity-volume proxy. It uses neither ACF's 43 quantities nor its
645-wire neighborhood expansion; it is not an ACF ablation.

P2Rank 2.5.1 is executed locally with a pinned binary and Java environment.
Ranking metrics use its native residue probability. Its shipped binary output
is the native pocket assignment. PocketMiner is rebuilt and checked on its
published self-test, where the reconstructed score is \PmSelfAuc{} versus the
reported \PmSelfPublished{}. The pLM-NN baseline uses the benchmark's ESM2-3B
embedding and trained head; reconstruction fidelity to the released example is
\PlmFidelity{} mean cosine similarity. A logistic regression on the same
\NTabWires{} continuous wires, standardized on the fit half, tests the
readout on the training selection half. For post-hoc external evaluation,
standardization and logistic coefficients were refitted on all 770 training
chains, retaining the training-derived propensity and the logistic model's
training-selected gate (18\,\AA, weight 0.5); the deployed ACF retains its
18\,\AA\ gate with weight 1. Both models scored the same residues. Set A and
Sets B+C were evaluated separately using 4,000 paired chain-bootstrap
resamples (seed 20260929) and unadjusted 95\% intervals. This supplementary
comparison was not preregistered.

The counting field's deployed binary call is the top
\TabOperatingQ{} fraction of residues per chain, selected on the training fold.
To remove an operating-point asymmetry, we additionally call the top
\MatchPTwoQ{} fraction of residues in each chain for both the field and P2Rank. Applying the same
grid search to P2Rank's training-fold predictions also selects
\MatchPTwoQ{}, so the common-budget and separately training-selected rules
coincide. Threshold curves from 2\% to 40\% are descriptive robustness analyses;
they do not change the deployed operating point. P2Rank's test-fold-best
threshold is shown only as a descriptive comparison in Fig.~\ref{fig:matched}.

\subsection{External validation}

The first external set (Set A) contains \ExtN{} apo--holo units and
\ExtNPositives{} positive residues from depositions released after
\ExtCutoff{}, the newest structure date in CryptoBench. It retains one unit per
UniRef50 cluster and shares no cluster with either CryptoBench partition.
Coordinates, labels, inclusion decisions, and hashes were frozen before
scoring. A committed analysis plan names three co-primary paired ROC-AUC
comparisons---the deployed field against P2Rank, pLM-NN, and PocketMiner---and
applies a Bonferroni correction. Binary comparisons use the same matched
top-\ExtQPct{}\% budget as the official-fold robustness analysis.

A second external read (Sets B+C) tests whether the training-fold lift from four
additional feature families transfers. These sets contain \BcN{} scorable
single-particle cryo-EM apo--holo units, are disjoint at UniRef50 from
CryptoBench, Set A, and each other, and were frozen before the additional
families were developed. One prespecified unit was excluded before scoring
because its three-character chain identifier could not be represented by the
legacy PDB writer. After the expanded 1269-column field was fixed, a committed
plan pinned four co-primary mean paired ROC-AUC comparisons: expanded field
minus the deployed 645-wire field, pLM-NN, P2Rank, and PocketMiner. All five
methods were evaluated on the same residue universe; no unit or residue was
lost to the intersection. Bonferroni-adjusted intervals control the four
comparisons. The repository commit graph is part of the chronology evidence
for both external reads and must accompany the archival submission.

\subsection{Statistical analysis}

Metrics are computed per unit and averaged across units. We report both
ROC-AUC and PR-AUC because precision--recall summaries are informative under
class imbalance.\cite{saito2015precision} For the official-fold and external
comparisons and the internal logistic comparison, intervals for paired
differences use a paired nonparametric bootstrap over units.\cite{efron1993bootstrap}
The main evaluations use \NBoot{} resamples; the internal and post-hoc
external logistic comparisons use 4,000. Undefined per-unit metrics are excluded from the
corresponding paired comparison; they are never imputed as zero.
``Resolved'' denotes a paired interval excluding zero, with multiplicity
adjustment where comparisons are co-primary. The mean paired ROC-AUC remains
the primary estimand. A 20\% trimmed mean that was selected on the training
partition is reported as exploratory because \EndReadsBefore{} indexed reads
of the official fold, \EndLineageBefore{} from the present architecture
lineage, preceded its registration.

Structural-family intervals are instead \(\bar d\pm1.96\,s_d/\sqrt{S}\),
where \(\bar d\) and \(s_d\) are the mean and sample standard deviation of
paired ROC-AUC gains across \(S=\FamAddOf\) half-splits. These descriptive
intervals ignore dependence from reused data and support no confirmatory
significance or population-coverage claims.

\subsection{Training-only structural-family probes}

To test what information the deployed representation omits, we appended
predefined feature families while holding the deployed pair bank, gate, ridge,
fan-out cap, and quantization fixed. Mean per-unit ROC-AUC lift over the frozen
deployed detector was measured on \FamAddOf{} UniProt accession-disjoint
halvings of the CryptoBench training fold (seeds 20260725--20260736).
Each family was compared with a size-matched control
that added the same number of tables over already deployed wires. For families
that read additional atom-level data, a second control permuted rows within
each chain, preserving column marginals while destroying residue assignment.
These probes are exploratory analyses of the representation; they do not
use the official or external folds for candidate selection or change the
deployed detector.

\section{Results}

\subsection{Official-fold ranking and strong baselines}

\begin{figure}[t]
\centering
\includegraphics[width=\textwidth]{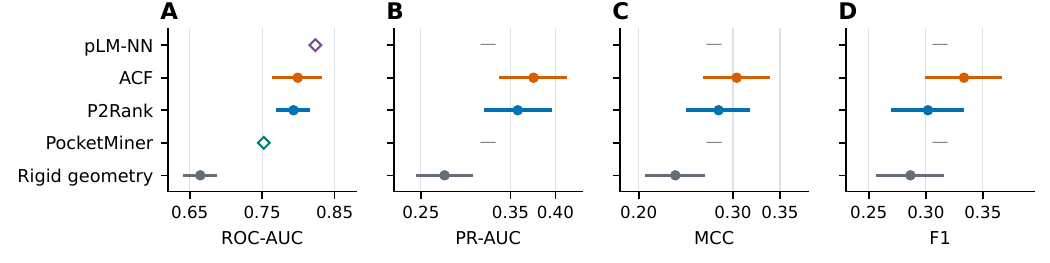}
\caption{Official-fold means (192 units). Filled points: means with 95\% bootstrap intervals; open diamonds: reported means without marginal intervals; dashes: metrics not shown. MCC/F1 use method-specific calls (ACF: top 9\%; P2Rank: native pockets). Axes differ. The paired ACF--P2Rank ROC-AUC difference remains unresolved.}
\label{fig:official}
\end{figure}

Table~\ref{tab:official} and Fig.~\ref{fig:official} summarize the
official-fold results.
The counting field reached ROC-AUC \TabAuc{} \TabAucCI{}, PR-AUC
\TabPr{} \TabPrCI{}, MCC \TabMcc{} \TabMccCI{}, and F1
\TabFOne{} \TabFOneCI{} on the official fold. P2Rank reached
\PtwoRAuc{}, \PtwoRPr{}, \PtwoRMcc{}, and \PtwoRFOne{}, respectively.
The paired mean ROC-AUC difference was \TabAucD{} \TabAucDCI{}
(\(p=\TabAucP{}\)); PR-AUC was likewise unresolved
(\TabPrD{} \TabPrDCI{}). The field was ahead on \EndWins{} chains and behind
on \EndLosses{}, with a median difference of \EndMedianDelta{}. The exploratory
20\% trimmed mean was \PreRegTestDelta{} \PreRegTestCI{}, but it cannot repair
the prior exposure of this fold and is not the primary claim.

\begin{table}[t]
\tbl{Official-fold results. The first three rows report all four metrics;
the last two report ROC-AUC for the reconstructed baselines.}
{\begin{minipage}{\textwidth}
\centering
\begin{tabular}{@{}lcccc@{}}
\toprule
Method & ROC-AUC & PR-AUC & MCC & F1\\
\midrule
Counting field & \TabAuc{} & \TabPr{} & \TabMcc{} & \TabFOne{}\\
P2Rank 2.5.1 & \PtwoRAuc{} & \PtwoRPr{} & \PtwoRMcc{} & \PtwoRFOne{}\\
Rigid geometry & \GeoAuc{} & \GeoPr{} & \GeoMcc{} & \GeoFOne{}\\
PocketMiner & \PmLevel{} & --- & --- & ---\\
pLM-NN & \PlmLevel{} & --- & --- & ---\\
\bottomrule
\end{tabular}
\end{minipage}}
\label{tab:official}
\end{table}

The stronger baselines bound the accuracy claim. The field-minus-pLM-NN
ROC-AUC difference was \PlmAucDelta{} \PlmAucCI{}; the field scored lower
on \PlmLoss{} of \NUnits{} chains. At a common calling budget, its F1 and MCC differences were
\PlmBudgetFDelta{} \PlmBudgetFCI{} and \PlmBudgetMDelta{}
\PlmBudgetMCI{}. pLM-NN remained ahead at all \CvN{} calling fractions. The
field exceeded PocketMiner by \PmAucOurs{} \PmAucCIOurs{}, but P2Rank also
exceeded PocketMiner by \PmAucPtwoR{} \PmAucCIPtwoR{}. Differences in label definitions may contribute to the PocketMiner gap;
these comparisons do not isolate its cause or establish general methodological
superiority.

On the training selection half, the same-wire logistic model reached ROC-AUC
\RdLogitAuc{}; the field-minus-logistic difference was
\RdLogitDelta{} \RdLogitCI{}. This training-fold comparison does not establish an accuracy advantage
for the table architecture.

\subsection{The F1 conclusion depends on the operating convention}

\begin{figure}[t]
\centering
\includegraphics[width=\textwidth]{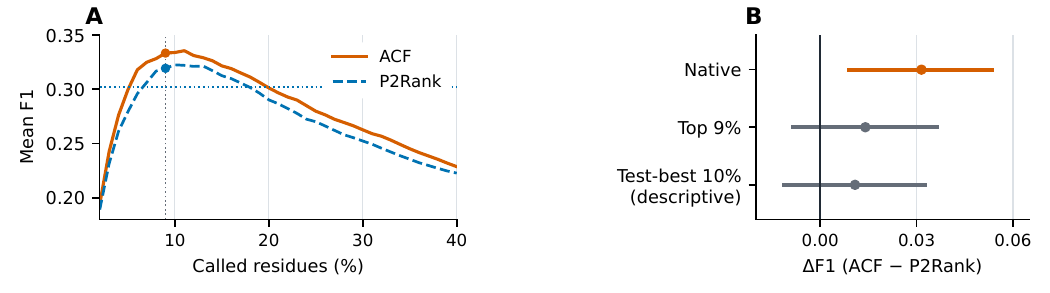}
\caption{Calling-budget sensitivity. (A) F1 curves; dotted lines mark training-selected 9\% calls (vertical) and P2Rank native-pocket F1 (horizontal). (B) Paired differences with 95\% intervals: native rules, matched top-9\%, and ACF at 9\% versus P2Rank's test-fold-best 10\% (descriptive). Matched-budget F1 is unresolved.}
\label{fig:matched}
\end{figure}

Fig.~\ref{fig:matched} compares the shipped and matched operating
conventions.
Under shipped outputs, counting-field F1 exceeded P2Rank by
\MatchNativeDelta{}. This compares the field's top-fraction rule
(\MatchOurQ{} of residues) with
P2Rank's native pocket assignment, which called \FullCallRatio{} times as many
residues. When both methods called the top \MatchPTwoQ{} fraction of residues
in each chain, their
mean F1 values were \MatchBOurs{} and \MatchBPTwo{}, and the paired difference
was \MatchBDelta{} \MatchBCI{} (\(p=\MatchBP{}\)). MCC was also unresolved
under the common budget (\FullBudgetMCCDelta{} \FullBudgetMCCCI{}). These
results do not support a convention-independent F1 advantage.

Across all \CvN{} thresholds from \CvLowPct{}\% to \CvHighPct{}\%, the field's
F1 remained directionally above P2Rank's, although pointwise intervals excluded
zero at only \CvPtwoRFExcl{} thresholds. pLM-NN remained above the field
throughout, with intervals excluding zero at all \CvN{} thresholds. This
robustness curve supports a directional comparison but does not create
\CvN{} independent confirmatory tests.

\subsection{First locked external validation}

Table~\ref{tab:external} and Fig.~\ref{fig:setbc}A report the predeclared external analysis. The field
reached mean ROC-AUC \ExtLevelOurs{}. Its paired advantage over P2Rank was \ExtAucPtwoR{} \ExtAucCIPtwoR{}, with a
Bonferroni-adjusted interval of \ExtAucBonfPtwoR{}; it was ahead on
\ExtWinPtwoR{} of \ExtN{} units. This is the paper's principal confirmatory
accuracy result.

\begin{table}[t]
\tbl{Locked external validation on \ExtN{} apo--holo units. Differences are
counting field minus comparator in mean paired ROC-AUC; intervals are
unadjusted. Bonferroni-adjusted intervals appear in Fig.~\ref{fig:setbc}A.}
{\begin{minipage}{\textwidth}
\centering
\begin{tabular}{@{}lccc@{}}
\toprule
Comparator & Comparator AUC & Difference & 95\% paired CI\\
\midrule
P2Rank & \ExtLevelPtwoR{} & \ExtAucPtwoR{} & \ExtAucCIPtwoR{}\\
pLM-NN & \ExtLevelPlm{} & \ExtAucPlm{} & \ExtAucCIPlm{}\\
PocketMiner & \ExtLevelPm{} & \ExtAucPm{} & \ExtAucCIPm{}\\
\bottomrule
\end{tabular}
\end{minipage}}
\label{tab:external}
\end{table}

The point-estimate deficit to pLM-NN persisted externally
(\ExtAucPlm{} \ExtAucCIPlm{}), although its multiplicity-adjusted interval
\ExtAucBonfPlm{} included zero. The advantage over PocketMiner was also
replicated (\ExtAucPm{} \ExtAucCIPm{}). At the matched top-\ExtQPct{}\%
budget, external F1 against P2Rank was unresolved
(\ExtBudgetFDelta{} \ExtBudgetFCI{}), agreeing
with the official-fold matched-threshold analysis. P2Rank abstained on
\ExtAbstain{} external unit; ranking analysis retained its residue scores,
whereas undefined threshold metrics followed the declared missing-value rule.

\subsection{Second external read of the expanded representation}

\begin{figure}[t]
\centering
\includegraphics[width=\textwidth]{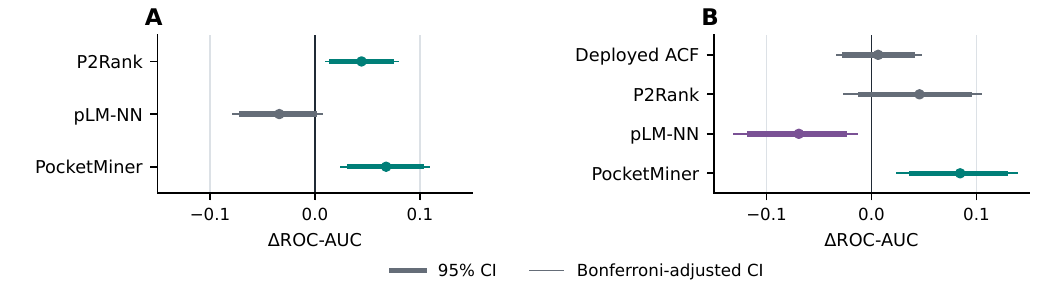}
\caption{Locked external evaluations. (A) Set A, 57 units: deployed ACF minus comparator. (B) Sets B+C, 45 cryo-EM units: expanded ACF minus comparator. Points are paired mean ROC-AUC differences; thick/thin lines denote 95\%/Bonferroni-adjusted intervals (three comparisons in A, four in B). Positive favors the named ACF variant. Expanded-versus-deployed improvement is unresolved; evaluations are not pooled.}
\label{fig:setbc}
\end{figure}

The expanded field reached mean per-unit ROC-AUC \BcLevelGeo{} on the \BcN{}
cryo-EM units, compared with \BcLevelOurs{} for the deployed field,
\BcLevelPtwoR{} for P2Rank, \BcLevelPm{} for PocketMiner, and \BcLevelPlm{} for
pLM-NN (Fig.~\ref{fig:setbc}B). The primary expanded-minus-deployed difference
was \BcOurs{} \BcOursCI{}, ahead on \BcOursWin{} units and behind on
\BcOursLoss{}. Its interval contains zero and the training-fold estimate
\BcOursPred{}; the read therefore does not confirm that the four added families
improve the deployed representation.

Against P2Rank, the expanded field differed by \BcPtwoR{} \BcPtwoRCI{} and
remained unresolved after multiplicity correction (\BcPtwoRBonf{}). It exceeded
PocketMiner by \BcPm{} \BcPmCI{}, with the adjusted interval \BcPmBonf{}
excluding zero. Conversely, pLM-NN remained ahead: the expanded-minus-pLM-NN
difference was \BcPlm{} \BcPlmCI{} (adjusted \BcPlmBonf{}). Thus, the expanded representation transfers signal relative to PocketMiner,
without a resolved gain over the deployed field or P2Rank.

In the post-hoc comparison, the deployed 645-wire ACF and logistic regression
achieved mean ROC-AUCs of 0.8411 and 0.8257 on Set A (57 units;
ACF minus logistic: $+0.0154$, 95\% CI [$-0.0012$, $+0.0347$]) and
0.7917 and 0.7968 on Sets B+C (45 units; $-0.0051$, 95\% CI
[$-0.0415$, $+0.0333$]), respectively. Neither interval excludes zero;
these results establish neither ACF superiority nor statistical equivalence.

\subsection{Discarded atom-level structure carries training-fold signal}

The controlled family ladder distinguishes new structural information from
re-encoding. A chemistry family that mainly resolves residue-identity
collisions in the existing physicochemical wires changed mean ROC-AUC by only
\FamChem{} \FamChemCI{}. In contrast, atom-level backbone descriptors improved
ROC-AUC by \FamBkb{} \FamBkbCI{}, side-chain conformational descriptors by
\FamSide{} \FamSideCI{}, and void-connectivity descriptors by \FamVoid{}
\FamVoidCI{}. The combined backbone--side-chain family yielded \FamConf{}
\FamConfCI{}, and adding void geometry produced \FamGeom{} \FamGeomCI{}.
All three corresponding size-matched controls remained near zero; within-chain
row permutation yielded negative gains for the backbone, side-chain, and void families.

The same screen identified a fourth source outside the deployed coordinate
summary. Within-chain ranks and ratios derived from deposited temperature
factors, occupancies, and alternate locations produced a \FamDisp{}
\FamDispCI{} lift, driven primarily by temperature-factor quantities
(\FamDispB{} \FamDispBCI{}) rather than alternate-conformer quantities
(\FamDispAlt{} \FamDispAltCI{}). These training-fold results support a
specific representation-level interpretation: useful information remains in
atom positions, empty-space connectivity, and structure-relative displacement
records that are lost when each residue is reduced to a centroid and a contact
neighborhood. They do not establish an accuracy gain for a revised detector on
held-out data.

The repository contains the full family ladder and controls. At the manuscript
snapshot, the expanded field's only held-out evaluation was Sets B+C;
neither the official fold nor Set A had been scored with it. Later probes on
the spent official fold remain exploratory.

\subsection{Architecture sensitivity, auditability, and cost}

Sensitivity models were fitted on the fit half and compared on the training
selection half; each row reports its best gate from the stated training-only
search, so the sweep is not uniformly a fixed-gate ablation. Three, four, and five quantization levels yielded ROC-AUC
0.7995, 0.8045, and 0.7996. Fan-out caps 16, 32, and 64 yielded 0.8018, 0.8045,
and 0.8010. Ridge values \TabRidgeLowValue{}, \TabRidgeMidValue{}, and
\TabRidgeHighValue{} yielded \TabRidgeLow{}, \TabRidgeMid{}, and
\TabRidgeHigh{}. At four levels, within-chain ranking reached 0.8045 versus
0.7905 for pooled ranking. The selected architecture ranked first on all
\NSplitsCv{} CryptoBench training folds and all \NSplitsRh{}
accession-disjoint half-splits. These results concern predictive performance, not stability of individual
integer weights. On the selection half, the frozen field scored 0.7873
without the spatial gate and 0.8045 with it.
This is not an independently held-out estimate of the gate benefit.

\begin{figure}[t]
\centering
\includegraphics[width=\textwidth]{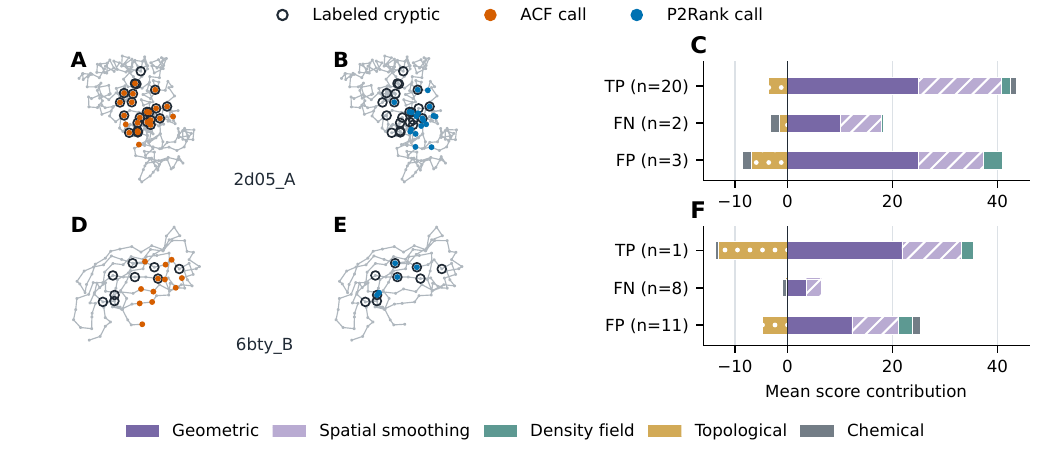}
\caption{Rule-selected disagreement cases: 2d05\_A (A--C) and 6bty\_B (D--F). A/D show ACF calls; B/E show P2Rank calls under deployed rules. Rings mark cryptic labels. C$\alpha$ projections preserve the coordinate-defined viewing direction and equal spatial scale within each case. C/F show mean ACF contributions for true positives (TP), false negatives (FN), and false positives (FP); n counts residues. Cases illustrate numerical traceability, not confirmatory evidence or causal mechanisms.}
\label{fig:cases}
\end{figure}

Exact recomputation of four residue scores from addresses, cell fractions,
multiplicities, and the spatial gate agreed with stored scores to within
\(5\times 10^{-16}\). Fig.~\ref{fig:cases} shows two rule-selected structural
disagreement cases. In these audited cases, the coarse geometry-family margin
between labeled hits and misses was \AudGeomMargin{}, and the gate added
\AudGateMargin{}. No coarse family separated true positives from false
positives by more than \AudGapToFP{}, so the decomposition identifies where
the score came from but is not a causal biological explanation.

On the matched \FairMachine{} environment, warm end-to-end median time was
\FairWarmOurs{}~s per chain for the field and \FairWarmPTwo{}~s for P2Rank:
ACF took \FairWarmRatio{} times as long and used
\FairCpuRatio{} times as much CPU time as P2Rank. Feature extraction consumed
\FairFeatures{}~s, or \FairFeatPct\% of the field runtime, while table scoring
consumed \FairScoring{}~s. In a cold one-chain invocation, P2Rank took \FairColdRatio{} times as long
as ACF because of JVM startup overhead. Thus, compact
model size, rather than faster served inference, is the supported efficiency
advantage (\FairModelOurs{}~MiB, \FairModelOursGz{}~MiB compressed, versus
\FairModelPTwo{}~MiB for the P2Rank model).

\section{Discussion}

\subsection{Positioning the contribution}

ACF's accuracy advantage over P2Rank is specific to Set A; the cryo-EM
comparison did not confirm gains from the expanded representation.
Same-wire logistic comparisons likewise do not establish a benefit of the
table readout. With pLM-NN stronger on the official fold and directionally
stronger externally, ACF's supported contributions are numerical traceability
and compact storage relative to P2Rank, not leading accuracy or faster warm
inference.

\subsection{Limitations}

The official test fold was inspected repeatedly: the manuscript snapshot
records three principal development rescoring rounds, 13 indexed reads
(12 from this architecture lineage and one from another), and 12 scored architectures. The subsequently updated repository reports 20
scored architectures; the original counts describe the snapshot, not the
current development history. Small official-fold margins are therefore descriptive rather than
confirmatory. Set A mitigates but does not eliminate this concern: it contains
only \ExtN{} units, excludes ambiguous intermediate-displacement cases through
its labeling guard band, and may be easier than CryptoBench. The second external
read contains only \BcN{} scorable units and consists entirely of
single-particle cryo-EM structures, so its wider intervals and modality shift
limit interpretation. The decline of all coordinate-reading methods relative
to the official fold is consistent with, but does not prove, reduced usable
local geometric signal in these models. Both external collections are now
spent and cannot confirm a future revision. Chronology must remain verifiable
from the full repository commit graph.

The existing sensitivity and family-addition analyses do not establish that
all 645 wires are essential, quantify their redundancy, or replace systematic
deletion ablations. Nonzero table weights indicate use, not independent
biological importance. Coordinate-perturbation robustness and stability of
individual weights across training splits and pairing seeds remain untested
here; bootstrap-seed stability addresses a different question. B-factor-derived
features may reflect structure quality or experimental refinement as well as
motion. Numerical auditability can support score inspection, but an actionable
experimental benefit has not been demonstrated.

The analysis excludes 38 multi-chain units and does not establish performance
at interfaces or in assemblies. Exact numerical decomposition is not the same
as mechanistic or causal explanation, particularly because hits and false
positives have similar coarse-family contributions. The target is a benchmark
cryptic-residue label, not ligand identity, binding affinity, druggability,
free-energy change, or biological efficacy. Broader locked datasets, a frozen
chain-aware mapping, and prospective experimental or simulation follow-up are
needed.

\section{Reproducibility}

Code, analysis protocols, and result artifacts are available at
\url{https://github.com/shaneraphel/geoaudit-cryptobench}.
The original results, macros, and deployed configuration are pinned to
\href{https://github.com/shaneraphel/geoaudit-cryptobench/tree/5d1726a067495f0e3fbbaf8af17652f8248f1d25}{commit \texttt{5d1726a}}.
The artifact \path{data/cryptobench_apo/TABLE_FIELD.json} records wire names,
pairs, counts, weights, propensity, operating point, and code/manifest hashes.
The manuscript uses \path{paper/frozen_numbers.tex} from this snapshot.
Defining modules are
\path{wide_descriptors.py}, \path{table_bank.py}, and \path{table_field.py}
under \path{src/pocket_bench/methods/}.
The post-hoc logistic protocol and results are in the
\href{https://github.com/shaneraphel/geoaudit-cryptobench-scaleup/tree/6c5a6400ad1c5171ccd88ea77c1b41c646bcf566}{companion repository (commit \texttt{6c5a640})}.
Pinned versions distinguish manuscript artifacts from later development.

\texttt{make recompute} regenerates metrics from archived predictions;
\texttt{make verify} checks artifacts. Inference and compilation require the
frozen inputs, caches, baseline binaries/weights, and environment.
\texttt{make extorder} checks lock/read ancestry in the retained Git history;
this supports chronology but cannot exclude unrecorded reads.

Runtime measurements used Apple M4, macOS 26.2 (arm64), Python 3.12.8,
NumPy 2.5.1/Accelerate, OpenJDK 17, and P2Rank 2.5.1. The record separates
warm/cold wall time, CPU time, threading, extraction, scoring, and peak RSS
(\FairRSS{}~MiB). These results are environment-specific; artifact checks
are not experimental reruns. This work received no funding and used only
public data (sources in Methods). X.W. is employed by Pfizer; the authors
declare no competing interests related to this work.

\section{Conclusion}

ACF provides compact, numerically traceable structure-based prediction without
homology or template searches at inference. On Set A, its mean paired ROC-AUC
advantage over P2Rank was 0.0443 (adjusted 95\% CI [0.0102, 0.0792]).
Official-fold ROC-AUC and matched-budget F1 differences against P2Rank remained
unresolved; cryo-EM evaluation did not confirm gains from added features.
These findings support an auditable supervised predictor, without establishing
lookup-table superiority or a causal explanation of pocket opening.

\bibliographystyle{ws-procs11x85}
\bibliography{references_psb}

\end{document}


\maketitle


\appendix
\section{Appendix A: every local quantity and every wire}
\label{app:wires}

This appendix is the definition of the input contract. The Methods section of
the main manuscript describes the field in terms of \(43\) local quantities
read under a fixed family of neighbourhood statistics; what follows states each
of the \(43\), the neighbourhood it is read over and its value at the boundary,
and then the rule that expands them into the \NTabWires{} wires. It is
generated from the modules, so it cannot describe a version of the code that is
not the one that produced the numbers.

\paragraph{Conventions.} \(c_i\) is the centroid of the heavy atoms of residue \(i\) and \(N(i)\) its contacts, meaning residues whose centroid lies within \(10\)~\AA. Neighbourhoods never cross a chain: a residue's neighbourhood is computed inside its own chain, which is what allows one chain to be scored without reference to any other. \(\hat n_i\) is the outward radial unit vector from the chain centroid, \(R_g\) the radius of gyration, \(s_i\) the sequence position, and \(a_i\) the number of heavy atoms. Hydrogens are discarded and a ligand-leak guard rejects any non-solvent heteroatom before any of this is computed.

\paragraph{The probe field.} Quantities in G1 and \texttt{rotor\_gain} are read off a free-space grid: points on a \(1.5\)~\AA\ lattice at least \(2.6\)~\AA\ from every atom, capped at \(6000\) points. The buriedness \(\beta(p)\) of such a point is the fraction of \(30\) Fibonacci-spaced directions along which an atom lies within \(11\)~\AA\ of the ray, at perpendicular tolerance \(1.8\)~\AA.

\paragraph{The ultrametric ball partition.} \(\rho_a\) is the coordination number of atom \(a\) within \(8\)~\AA. Contact edges are deflated to \(\|c_a - c_b\| (\rho_{\max}/\min(\rho_a,\rho_b))^{\gamma}\) with \(\gamma = 2\), and the balls at radius \(\tau\) are the components of the minimum spanning tree restricted to edges of weight at most \(\tau\), which is exactly the closed-ball partition of the minimax ultrametric. \(\tau\) is chosen by the valuation threshold of the tree-weight spectrum, not tuned. Note that the \texttt{valuation} descriptor below uses its own exponent \(0.5\), not \(\gamma\).

\subsection{The \NAlgFeatures{} algebraic and topological quantities}

The grouping is not cosmetic: each group is one dense quaternary table in the cascaded compiler, so no group may exceed six digits.

\paragraph{G1 surface exposure (6).}
\begin{description}
  \item[\texttt{bur}] mean over probe points $p$ attributed to $i$ of the buriedness $\beta(p)$, the fraction of $30$ Fibonacci directions blocked by an atom within $11$~\AA\ of the ray, at perpendicular tolerance $1.8$~\AA. \emph{Read over}: free-grid probe points within $6$~\AA\ of any heavy atom of $i$. \emph{At the boundary}: $0$ when no probe point is attributed to $i$.
  \item[\texttt{void}] $\log\!\left(1 + \#\{p : \beta(p) \ge 0.55\}\right)$. \emph{Read over}: the same attributed probe points. \emph{At the boundary}: $\log 1 = 0$ when none of them is that buried.
  \item[\texttt{probe\_max}] $\max_p \beta(p)$. \emph{Read over}: the same attributed probe points. \emph{At the boundary}: $0$ when no probe point is attributed to $i$.
  \item[\texttt{protrusion}] $\left\| \operatorname{mean}_{j \in N(i)} c_j - c_i \right\| / R$, the offset of the neighbourhood's centroid from the residue. \emph{Read over}: residues within $R = 10$~\AA. \emph{At the boundary}: $0$ when $i$ has no contact.
  \item[\texttt{exposure}] $\#\{p\} / a_i$, probe points per heavy atom of the residue. \emph{Read over}: probe points within $5$~\AA\ of any heavy atom of $i$. \emph{At the boundary}: $0$ when none; $a_i \ge 1$ by construction.
  \item[\texttt{concavity}] $\operatorname{mean}_{j \in N(i)} \widehat{(c_j - c_i)} \cdot \hat n_i$ with $\hat n_i$ the outward radial unit vector from the chain centroid. \emph{Read over}: residues within $10$~\AA. \emph{At the boundary}: $0$ when $i$ has no contact.
\end{description}

\paragraph{G2 local spectral geometry (6).}
\begin{description}
  \item[\texttt{fiedler}] $\lambda_2$ of the combinatorial Laplacian $L = D - A$ of the lining $G[S_i]$, $S_i = \{i\} \cup N(i)$. \emph{Read over}: induced subgraph on $i$ and its contacts, capped at the $48$ nearest. \emph{At the boundary}: $0$ when $i$ has no contact or $|S_i| \le 1$.
  \item[\texttt{lap\_max}] $\lambda_{\max}$ of the same $L$. \emph{Read over}: the same lining. \emph{At the boundary}: $0$ when $i$ has no contact.
  \item[\texttt{spec\_gap}] $\lambda_2 / \max(\lambda_{\max}, 10^{-9})$. \emph{Read over}: the same lining. \emph{At the boundary}: the floor in the denominator is what makes it defined at $\lambda_{\max} = 0$.
  \item[\texttt{mean\_degree}] $2E / |S_i|$ of the lining. \emph{Read over}: the same lining. \emph{At the boundary}: $0$ when $i$ has no contact.
  \item[\texttt{betti0}] number of connected components of $G[S_i \setminus \{i\}]$ --- how many disjoint walls face the residue. \emph{Read over}: the lining re-thresholded at $7$~\AA, because at $10$~\AA\ the lining of a globular protein is connected and this count is the constant $1$. \emph{At the boundary}: $0$ when the lining is empty.
  \item[\texttt{betti1}] cycle rank $E - V + C$ of the same graph, an integer. \emph{Read over}: the same $7$~\AA\ lining. \emph{At the boundary}: $0$ when the lining is empty.
\end{description}

\paragraph{G3 density field calculus (6).}
\begin{description}
  \item[\texttt{rho}] mean over the residue's heavy atoms of the atomic packing density $\rho$, the coordination number within $8$~\AA. \emph{Read over}: the residue's own atoms. \emph{At the boundary}: $\rho \ge 1$ by construction.
  \item[\texttt{grad\_rho}] $\|g_i\|$ where $g_i$ solves the normal equations $A_i g_i = b_i$ of the first-order Taylor fit, $A_i = \sum_j (c_j - c_i)(c_j - c_i)^\top$, $b_i = \sum_j (\rho_j - \rho_i)(c_j - c_i)$. \emph{Read over}: residues within $10$~\AA. \emph{At the boundary}: $g_i = 0$ when $A_i$ is singular, which is a neighbourhood spanning fewer than three dimensions.
  \item[\texttt{grad\_radial}] $g_i \cdot \hat n_i$, the outward component of the same gradient. \emph{Read over}: the same neighbourhood. \emph{At the boundary}: $0$ with $g_i$.
  \item[\texttt{lap\_rho}] $\operatorname{mean}_{j \in N(i)} \rho_j - \rho_i$, the graph Laplacian of the density field. \emph{Read over}: residues within $10$~\AA. \emph{At the boundary}: $0$ when $i$ has no contact.
  \item[\texttt{var\_rho}] $\max\!\left(\operatorname{mean}_j \rho_j^2 - (\operatorname{mean}_j \rho_j)^2,\, 0\right)$. \emph{Read over}: the same neighbourhood. \emph{At the boundary}: clipped at zero so rounding cannot produce a negative variance.
  \item[\texttt{rho\_rank\_ball}] rank of $\rho_i$ among the residues of $i$'s ultrametric ball, scaled to $[0,1]$. \emph{Read over}: the ball, not a radius. \emph{At the boundary}: $0.5$ when the ball holds one residue, which is the value that asserts nothing.
\end{description}

\paragraph{G4 ultrametric structure (6).}
\begin{description}
  \item[\texttt{loose}] mean over the residue's atoms of $\bar\rho_{B(a)} / \bar\rho$, the atom's ball density against the global mean. \emph{Read over}: the residue's atoms and their balls. \emph{At the boundary}: $\bar\rho$ is floored at $1$.
  \item[\texttt{ball\_size}] $\log(1 + |B(i)|)$ in atoms. \emph{Read over}: the ball. \emph{At the boundary}: $\log 1 = 0$ for an empty ball, which cannot occur.
  \item[\texttt{ball\_radius}] $\tau \cdot \bar\rho / \bar\rho_{B(i)}$, the ball's cut radius deflated by its own density. \emph{Read over}: the ball. \emph{At the boundary}: the denominator is floored at $10^{-9}$.
  \item[\texttt{ball\_interface}] fraction of $i$'s contacts lying in a different ball. \emph{Read over}: residues within $10$~\AA. \emph{At the boundary}: $0$ when $i$ has no contact.
  \item[\texttt{valuation}] $\max_{j \in N(i)} \|c_j - c_i\| \left(\rho_{\max} / \min(\rho_i, \rho_j)\right)^{0.5} \big/ \tau$: the longest density-deflated edge the residue carries, in units of the global ball radius. \emph{Read over}: residues within $10$~\AA. \emph{At the boundary}: $0$ when $i$ has no contact; the deflator's denominator is floored.
  \item[\texttt{rotor\_gain}] largest increase in probe buriedness produced by rotating one ultrametric ball rigidly by $\theta = 0.20$~rad about its branch axis. \emph{Read over}: the $8$ largest balls of at least $8$ atoms that are not the whole structure; probes within $19$~\AA\ of the hinge centre. \emph{At the boundary}: $0$ when no ball qualifies, which is the honest reading that the structure has no hinge at this scale.
\end{description}

\paragraph{G5 curvature and anisotropy (6).}
\begin{description}
  \item[\texttt{anisotropy}] $(\lambda_1 - \lambda_3)/\operatorname{tr} M$ for $M_i = \operatorname{mean}_{j \in N(i)} (c_j - c_i)(c_j - c_i)^\top$ with $\lambda_1 \ge \lambda_2 \ge \lambda_3$. \emph{Read over}: residues within $10$~\AA. \emph{At the boundary}: $\operatorname{tr} M$ floored at $10^{-12}$.
  \item[\texttt{planarity}] $(\lambda_2 - \lambda_3)/\lambda_1$. \emph{Read over}: the same tensor. \emph{At the boundary}: $\lambda_1$ floored at $10^{-12}$.
  \item[\texttt{sphericity}] $\lambda_3/\lambda_1$. \emph{Read over}: the same tensor. \emph{At the boundary}: $\lambda_1$ floored at $10^{-12}$.
  \item[\texttt{normal\_div}] $\operatorname{mean}_{j} (\hat n_j - \hat n_i) \cdot (c_j - c_i)$, the discrete divergence of the outward normal field. \emph{Read over}: residues within $10$~\AA. \emph{At the boundary}: $0$ when $i$ has no contact.
  \item[\texttt{angle\_deficit}] $1 - \operatorname{mean}_j \angle(\hat n_i, \hat n_j) / \pi$, bounded in $[0,1]$ by construction. \emph{Read over}: residues within $10$~\AA. \emph{At the boundary}: $1$ when $i$ has no contact, the value a flat isolated patch takes.
  \item[\texttt{hess\_bur}] $\operatorname{mean}_{j} \mathrm{bur}_j - \mathrm{bur}_i$, the graph Laplacian of the buriedness field. \emph{Read over}: residues within $10$~\AA. \emph{At the boundary}: $0$ when $i$ has no contact.
\end{description}

\paragraph{G6 global position (5).}
\begin{description}
  \item[\texttt{depth}] $\|c_i - \bar c\| / R_g$ against the chain centroid and radius of gyration. \emph{Read over}: the whole chain. \emph{At the boundary}: $R_g$ floored at $1$.
  \item[\texttt{coordination}] $|N(i)|$, an integer. \emph{Read over}: residues within $10$~\AA. \emph{At the boundary}: $0$ is attainable and meaningful.
  \item[\texttt{contact\_order}] $\operatorname{mean}_{j \in N(i)} |s_j - s_i| / n$, the mean sequence separation of the contacts normalised by chain length. \emph{Read over}: residues within $10$~\AA. \emph{At the boundary}: $0$ when $i$ has no contact.
  \item[\texttt{ball\_offset}] $\|c_i - \bar c_{B(i)}\| / R_g$, displacement from the residue's own ball centroid. \emph{Read over}: the ball. \emph{At the boundary}: $R_g$ floored at $1$.
  \item[\texttt{depth\_rank}] rank of $\mathrm{depth}$ within the chain, scaled to $[0,1]$. \emph{Read over}: the whole chain. \emph{At the boundary}: $0$ for a chain of one residue.
\end{description}

\subsection{The seven physicochemical constants and the propensity}

None of these is fitted to CryptoBench. The seven are published constants of the residue type; the eighth is a count.

\begin{description}
  \item[\texttt{kd}] Kyte--Doolittle hydropathy. \emph{Source}: Kyte and Doolittle, \emph{J.\ Mol.\ Biol.}\ \textbf{157}:105 (1982).
  \item[\texttt{volume}] side-chain volume in \AA$^3$. \emph{Source}: Zamyatnin, \emph{Prog.\ Biophys.\ Mol.\ Biol.}\ \textbf{24}:107 (1974).
  \item[\texttt{aromatic}] aromatic rings in the side chain: 1 for F, Y, H, 2 for W, 0 otherwise. \emph{Source}: chemical structure.
  \item[\texttt{charge}] formal charge at pH 7: $-1$ for D, E, $+1$ for K, R, $+0.1$ for H as partially protonated, 0 otherwise. \emph{Source}: chemical structure.
  \item[\texttt{hbd}] side-chain hydrogen-bond donors. \emph{Source}: chemical structure.
  \item[\texttt{hba}] side-chain hydrogen-bond acceptors. \emph{Source}: chemical structure.
  \item[\texttt{chi}] rotatable side-chain bonds ($\chi$ angles). \emph{Source}: chemical structure.
\end{description}

An unknown or non-standard residue takes the mean of the twenty standard values rather than a sentinel, so it is neutral in the rank order rather than extreme in it.

\paragraph{The propensity, and why it is training-only.} The forty-third quantity is \(\hat P(\text{cryptic} \mid \text{residue type})\), Laplace-smoothed by one pseudo-count per class,
\begin{equation}
  \pi_t = \frac{1 + \sum_{i : \mathrm{type}(i) = t} y_i}{2 + \#\{i : \mathrm{type}(i) = t\}},
\end{equation}
counted over the training partition only. It is a bincount, not a fit. The table is written into the compiled artifact and read back at inference, so a test residue never contributes a count to the table that scores it; a type absent from training receives the neutral prior \(1/2\) rather than an undefined ratio. This is the only one of the \(43\) that touches a label, and it is the reason the field is described as compiled on the training partition rather than as training-free.

\subsection{From \(43\) local quantities to \NTabWires{} wires}

Each local quantity \(x\) is read under five kinds of statistic over intra-chain neighbourhoods \(N_r(i)\) of fixed radius:
\begin{align}
  \text{identity} &: x_i, \\
  \text{mean} &: \textstyle \frac{1}{|N_r(i)|} \sum_{j \in N_r(i)} x_j, \\
  \text{dispersion} &: \Bigl( \textstyle \frac{1}{|N_r(i)|} \sum_j x_j^2 - \bigl(\frac{1}{|N_r(i)|}\sum_j x_j\bigr)^2 \Bigr)^{1/2}, \\
  \text{centred difference} &: x_i - \textstyle \frac{1}{|N_r(i)|}\sum_j x_j, \\
  \text{local rank} &: \textstyle \frac{1}{|N_r(i)|} \bigl|\{ j \in N_r(i) : x_j < x_i \}\bigr|.
\end{align}

The radii are not the same for every statistic, and the counts are what produce \NTabWires{}:

\begin{center}
\begin{tabular}{llr}
\toprule
statistic & radii (\AA) & groups \\
\midrule
identity & --- & 1 \\
mean & 6, 10, 14, 20, 26 & 5 \\
dispersion & 6, 14, 20 & 3 \\
centred difference & 6, 14, 20 & 3 \\
local rank & 6, 14, 20 & 3 \\
\midrule
total & & 15 \\
\bottomrule
\end{tabular}
\end{center}

So \(35\) algebraic quantities, \(7\) constants and one propensity give \(43\) local quantities, and \(43 \times 15 = 645\) wires. The local rank is the only statistic in the set that is not a moment: it asks where a residue sits in the order of its neighbourhood rather than how far it is from the neighbourhood's centre, so it is unchanged by any monotone rescaling of the quantity, and it separates a residue that is marginally the most hydrophobic of its neighbours from one that is marginally the least --- a distinction the centred difference blurs into two small numbers of opposite sign.

\paragraph{Banding.} Each wire is then banded into a quaternary digit by its rank \emph{within its own chain}, ties sharing a mid-rank:
\begin{equation}
  d_i = \min\!\left( \left\lfloor \frac{r_i}{\max(n-1,1)} \, L \right\rfloor,\, L - 1 \right), \qquad L = 4 .
\end{equation}
Ranking inside the chain is what makes a wire mean the same thing in a 57-residue chain and a 307-residue one, and it removes any dependence on absolute units, so no constant has to be carried from the training fold in order to score a new structure.